**Evaluation of ChatGPT Family of Models for Biomedical Reasoning and Classification**

*Shan Chen,[1,2] *Yingya Li,[3] Sheng Lu,[4] Hoang Van,[3] Hugo JWL Aerts,[1,2,5] Guergana K. Savova,[3] Danielle S. Bitterman,[1,2]

[1.] Artificial Intelligence in Medicine (AIM) Program, Mass General Brigham, Harvard Medical School, Boston, MA

[2.] Department of Radiation Oncology, Brigham and Women's Hospital/Dana-Farber Cancer Institute, Boston, MA

[3.] Computational Health Informatics Program, Boston Children's Hospital, and Harvard Medical School, Boston, MA

[4.] Ubiquitous Knowledge Processing Lab, Technische Universität Darmstadt, Darmstadt, Germany

5. Radiology and Nuclear Medicine, GROW & CARIM, Maastricht University, The Netherlands

* indicates co-first authors

**ABSTRACT**

**Objective/Purpose:** Recent advances in large language models (LLMs) have shown impressive ability in biomedical question-answering, but have not been adequately investigated for more specific biomedical applications. This study investigates the performance of LLMs such as the ChatGPT family of models (GPT-3.5, GPT-4) in biomedical tasks beyond question-answering.

**Materials/Methods:** Because no patient data can be passed to the OpenAI API public interface, we evaluated model performance with 10,000+ samples as proxies for two fundamental tasks in the clinical domain – classification and reasoning. The first task is classifying whether statements of clinical and policy recommendations in scientific literature constitute health advice. The second task is causal relation detection from the biomedical literature. We used 20% of the dataset for prompt development under the settings of zero- and few-shot with and without chain of thought (CoT). The most effective prompt from each setting was evaluated on the remaining 80%. We compared LLMs with models using simple features (bag-of-words (BoW)) with logistic regression, and fine-tuned BioBERT models.

**Results:** Fine-tuning BioBERT yielded the best results for classification (F1 0.800-0.902) and reasoning (F1 0.851). Of the LLM approaches, few-shot CoT yielded the best results for classification (F1 0.671-0.770) and reasoning (F1 0.682), comparable to the BoW model (F1 0.602-0.753 and 0.675 for classification and reasoning, respectively). The total time needed to achieve the best LLM results was 78 hours, compared to 0.078 and 0.008 hours to develop the best-performing BioBERT and BoW models, respectively.

**Conclusions:** Despite the excitement around viral ChatGPT, we found that fine-tuning for two fundamental NLP tasks remained the best strategy. The simple BoW model performed on par with the most complex LLM prompting. Prompt engineering required significant investment.

**Funding: R01GM114355 (YL), R01LM013486 (HV, GS), Woods Foundation (SC, DB)**

# MAIN PAPER

## 1.0 Introduction and Background

The advancements in machine learning (ML) methods for natural language process (NLP), such as transformers[1] and reinforcement learning[2], in combination with abundant digital text and scaled-up hardware capabilities has led to many pretrained large language models (LLMs)—also referred to as foundation models. Coupling some of these LLMs with smart engineering gave the world the viral ChatGPT, which in turn popularized the technology and re-invigorated the artificial intelligence (AI)/artificial general intelligence (AGI) debate. Although most LLMs are trained as chatbots, some of the claims in the mainstream media go as far as stating that the LLMs are sentient, even able to solve tasks that previously required a high level of human expertise and specialized training. On the other hand, the scientific papers describing the LLMs are much more measured[3] outlining limitations: *"… Aside from intentional misuse, there are many domains where large language models should be deployed only with great care, or not at all. Examples include high-stakes domains such as medical diagnoses, classifying people based on protected characteristics, determining eligibility for credit, employment, or housing, generating political advertisements, and law enforcement."[3]* Therefore the scientific community bears the responsibility of understanding the LLMs' strengths and weaknesses, how their limitations and risks can be managed, and implications for our future.[4,5]

Medicine is one of the highest-stakes domains for LLMs. The excitement surrounding the LLMs has penetrated the biomedical and clinical communities motivating various early use-case evaluations. Two studies[6,7] evaluate ChatGPT on the US Medical Licensing Examination (USMLE) tests suggesting it holds passing scores. Zuccon and Koopman[8] investigate the effect of prompts on ChatGPT in answering complex health information questions. Chen et al.[9] evaluate the performance and robustness of ChatGPT in providing cancer treatment recommendations that align with National Comprehensive Cancer Network (NCCN) guidelines. Lyu et al.[10] research the feasibility of using ChatGPT to translate radiology reports into plain language for patients and healthcare providers. A paper by Google Research and DeepMind[11] presents experiments with Google's PaLM family of LLMs, suggesting the potential utility of LLMs in medicine but also revealing important limitations, reinforcing the importance of evaluation frameworks and methods development.

In parallel, the practical use of LLMs is limited by their huge size and computational requirements, limiting accessibility for most healthcare practices and researchers. Thus, researchers have been pursuing broader questions such as the utility of specialized clinical models, especially ones that are smaller and thus computationally affordable, in the LLM era. Lehman et al.[12] show that relatively small specialized clinical models substantially outperform bigger LLMs, even when fine-tuned on limited annotated data. In addition, they show that pretraining on clinical datasets allows for smaller, more parameter-efficient models that either match or outperform the much bigger computationally hungry LLMs. Wang et al.[13] focus on exploring ChatGPT robustness where a medical diagnosis dataset represents out-of-domain distributions. Results are consistent with Lehman at al.[12]

We set out to contribute to the growing understanding of LLMs in the biomedical domain, with a focus on practical end-use. NLP research on the clinical narrative within the Electronic Medical Records (EMR) has direct applications to translational science[14], clinical decision support,[15] and healthcare administration[16] in addition to direct patient care. Two fundamental NLP tasks to support these applications are classification (e.g. patient phenotyping) and reasoning (e.g. adverse events of medications). Thus, we aim to evaluate the state-of-the-art (SOTA) LLM performance on classification and reasoning tasks requiring understanding of contextual nuances. While numerous LLMs have been trained in recent years, most are proprietary and not publicly available for a local download (e.g., GPT-3, ChatGPT), which precludes their evaluation on clinical datasets containing Protected Health Information (PHI) data, even if the data are de-identified. Therefore, we work with proxy biomedical data. We evaluate LLMs within the constraints of the typical user to understand their real-world utility, using the OpenAI API[17] and LLMs that are computationally feasible for the IT capabilities of most hospitals and clinical practice.

## 2.0 Experimental Setup

### 2.1 Tasks and datasets

We examine the performance of LLMs on two fundamental tasks in the clinical domain – classification and reasoning.[1] Specifically, we select two open datasets annotated for health advice and causal language to test the ability of the models to classify and reason over medical literature findings and their implications for health-related practices.

- **Classification task: HealthAdvice**[18] is a dataset consisting of annotations of 10,000+ sentences extracted from abstracts and discussion/conclusion sections of medical research literature. The dataset adopts a multi-dimensional taxonomy and categorizes each sentence into "no advice", "weak advice", and "strong advice" to capture the occurrence and strength of clinical and policy recommendations. As health advice normally appears in either abstracts or discussion/conclusion sections and its language style may vary across different sections, the labels are further separated into 3 datasets: *advice in discussion sections, advice in unstructured abstracts, and advice in structured abstracts*.
- **Reasoning task: CausalRelation**[19] is a multi-label reasoning dataset with the goal to identify correlational and causal claims in the findings of medical research literature. The annotated corpus includes over 3,000 PubMed research conclusion sentences extracted from abstracts. Each sentence is labeled as "correlational", "conditional causal", "direct causal", and "no relationship" by its certainty and reasoning type.

Table 1(A) and 1(B) show the dataset distributions. We split the two datasets into development and test sets. The development set is a proportionate sample of 20% the original dataset, while the remaining 80% of the dataset is used as the test set. Final evaluation is performed on the test set.

---

[1] All codes, prompts and associate outputs can be find: https://github.com/shan23chen/HealthLLM_Eval

| Task | Label | Advice in discussion sections | Advice in unstructured abstracts | Advice in structured abstracts |
|---|---|---|---|---|
| Classification: HealthAdvice | Weak advice | 162 | 28 | 1482 |
| | Strong advice | 135 | 16 | 925 |
| | No advice | 3635 | 890 | 3575 |
| | Total | 3932 | 934 | 5982 |

*Table 1(A). HealthAdvice dataset distributions*

| Task | Label | Count |
|---|---|---|
| Reasoning: CausalRelation | Correlational | 998 |
| | Conditional causal | 213 |
| | Causal | 494 |
| | No relationship | 1356 |
| | Total | 3061 |

*Table 1(B). CausalRelation dataset distributions*

**2.1 Baseline models**
We compare the performance of LLMs with classic ML approaches and transformer-based pretrained language models. For the classic ML approach, we train logic regression fitted with Stochastic Gradient Descent (SGD), using bag-of-words (BoW) representations with tf-idf as the vectorization method. We fine-tune BioBERT models, given that BERT-based pre-trained language models[20], particularly BioBERT[21], have exhibited their efficacy on the aforementioned two tasks (Hyperparameter settings are in the Appendix Table A1). The baseline models are trained and fine-tuned on the full development set and tested on the test set. To further examine the effect of the amount of data on model performance, we trained and tested the BoW and BioBERT models with the 20%, 50%, and 100% of the development set. We track the time required to develop and evaluate the BoW and BioBERT models.

**2.2 ChatGPT Family of Models**
We evaluate GPT-4 and its predecessors, including GPT-3.5-Turbo(20B) and GPT-Davinci-003(175B)), on the two tasks. For GPT-3.5 models, we consider zero-shot, one-shot, and few-shot prompting with and without Chain-of-Thought (CoT). Given computational cost limits, for GPT-4, we consider zero-shot (the simplest prompting strategy mimicking an average user) and few-show with CoT (the most complex prompting strategy). CoT techniques explicitly outline the intermediate reasoning steps as prompts to LLMs to elicit multi-step reasoning behavior. [22]

To design a prompt, we follow the prompt structure applied in prior studies.[22] Fig 1 shows an example of the prompt templates for the classification task. For the zero-shot settings, five of the authors each independently developed two prompts. For the one- and few-shot settings without CoT, exemplars are chosen directly from the development set. For the one- and few-shot settings with CoT, the same five authors independently wrote CoT prompts for exemplars from each of the datasets. To evaluate model efficiency, we track the total time spent on designing prompts. Given the different number of classes in the health advice (3 classes) and causal language (4 classes) datasets, we apply 3- and 4-shot exemplars for the few-shot settings for the classification and reasoning tasks respectively. We use regular expressions to match the model output to labels in the datasets, and conduct validation to verify the accuracy of the regular expressions. To assess the model's performance, we compare the prediction results against the gold annotations in the datasets. Performance of the prompts was initially evaluated on the development set, and the best performing prompt was selected for the final evaluation on the test dataset. Performance across exemplars from different classes was also evaluated. The full set of evaluated prompts are included in the Appendix Table B1 and B2.

We use the OpenAI API to run the prediction and measure the performance of the models based on the averaged macro-F1 score on 4-fold cross validation on the test set. F1 score is a classic NLP metric representing the harmonic mean of recall/sensitivity and precision/positive predictive value. Macro-F1 score is computed using the arithmetic mean of all per-class F1 scores. For comparison of model efficiency, we track the time and cost for running the inference using the API (Appendix Table A2 a-c)

[Original Context] = Georgian public health specialists working in the HIV field should prioritize implementation of such interventions among HIV patients.
[Question] = Is this a 0) no advice, 1) weak advice or 2) strong advice?
[Answer] = 2) Strong advice.
[CoT Solution] =
1) The statement is a directive, so it is a strong advice.
2) The statement is specific and clear, so it is not a weak advice.
3) Therefore, the statement is a strong advice.
Answer: 2) Strong advice.

A) Zero-Shot

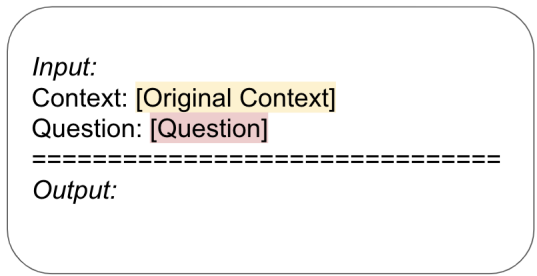

B) Zero-Shot CoT

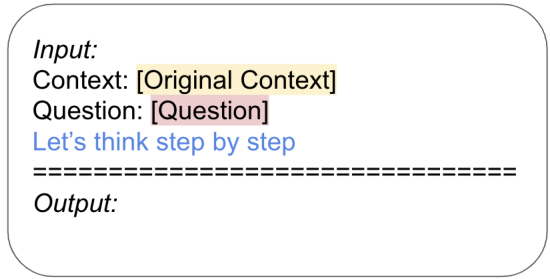

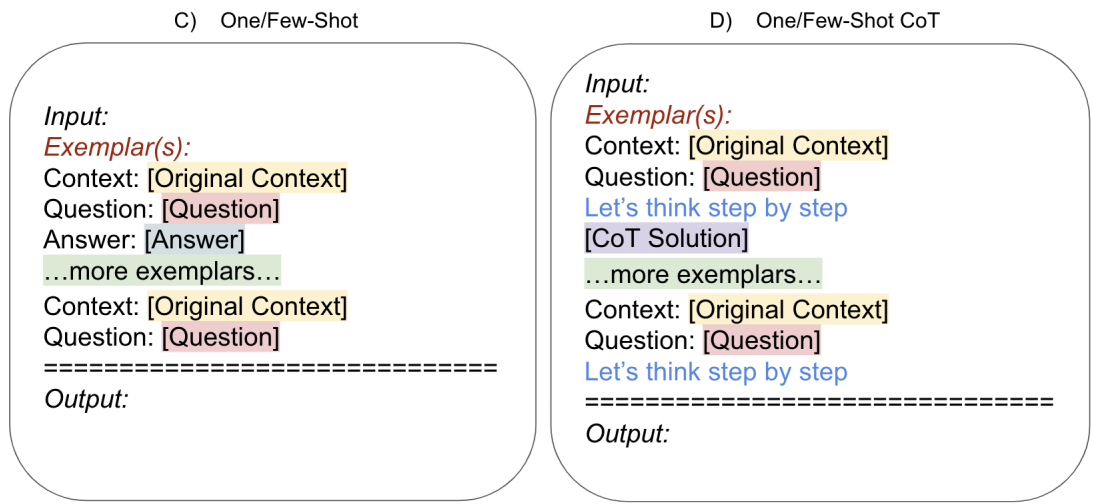

*Figure 1.* Examples of the prompt templates used for the classification and reasoning tasks. Prompt templates were created for each of the prompting settings evaluated: **(A)** zero-shot, **(B)** zero-shot with Chain of Thought (CoT), **(C)** one-shot and few-shot, and **(D)** one-shot and few-shot with CoT. "…more exemplars…" (highlighted in green) were added only for the few-shot with and without CoT settings. All evaluated prompts are in the Appendix. BoW = bag-of-words.

### 2.3 Smaller LLMs

The same settings (zero- and few-shot with and without CoT) were used to evaluate the performance of select smaller LLMs (less than 10B parameters) on the tasks, including GPT-J, GPT-JT, and Galactica[23]. GPT-J, built on EleutherAI's 6B parameter GPT-J-6B, is fine-tuned with 3.5 billion tokens. It performs very similarly to GPT-3 on various zero-shot downstream tasks. GPT-JT, a fork of GPT-J-6B, is fine-tuned on 3.53 billion tokens and has been shown to even outperform GPT-3 at some classification tasks. Galactica is trained on 48 million examples of scientific articles, websites, textbooks, lecture notes, and encyclopedias. Same evaluation procedure as for the GPT-family models is applied.

### 2.4 Statistical Analysis

Paired *t* tests are used to compare average macro-F1 score across tasks, and a 2-sided $p<0.05$ is considered statistically significant. Analyses are performed using python version 3.9.7 (Python Software Foundation).

# 3.0 Results

A summary of our findings is in Table 2, Fig 2, Fig 3 and 4 present multiple comparative analyses to provide context and facilitate interpretation of the outcomes.

| Model | Settings | Classification | | | Reasoning |
|---|---|---|---|---|---|
| | | Advice in discussion sections | Advice in unstructured abstracts | Advice in structured abstracts | Causal relation detection |
| | Random | **0.209** | **0.174** | 0.313 | 0.231 |
| GPT-J, GPT-JT, Galactica | Best-6.7B models[2] | **0.276** | **0.284** | 0.358 | **0.244**[3] |
| GPT-4 | zero-shot | 0.509 | 0.478 | 0.517 | 0.301 |
| | few-shot CoT | **0.648** | **0.712** | **0.770** | **0.682** |
| GPT3.5s (Davinci-003/ ChatGPT-Turbo) | zero-shot | 0.506 | 0.489 | 0.548 | 0.288 |
| | one-shot | **0.513** | **0.554** | **0.593** | 0.396 |
| | few-shot | 0.475 | 0.403 | 0.495 | **0.542** |
| | zero-shot CoT | 0.436 | 0.424 | 0.515 | 0.345 |
| | one-shot CoT | 0.430 | 0.410 | 0.552 | 0.553 |
| | few-shot CoT | **0.671** | **0.670** | **0.718** | **0.649** |
| BoW | 20% BoW | 0.551 | 0.566 | 0.701 | 0.465 |
| | 50% BoW | **0.602** | 0.609 | 0.735 | 0.581 |
| | 100% BoW | 0.593 | **0.640** | **0.753** | **0.675** |
| BioBERT | 20% FT BioBERT | 0.682 | 0.787 | 0.791 | 0.597 |
| | 50% FT BioBERT | 0.703 | 0.810 | 0.864 | 0.760 |
| | 100% FT BioBERT | **0.800** | **0.821** | **0.902** | **0.851** |

*Table 2. Summary of best macro-F1 results across datasets and models. FT = fine-tuning. Bolded text indicates best performance. Percentage for BoW and BioBERT indicates how much development set data was used for training in a supervised setting. "Random" reflects the uniform distribution across class labels. Comprehensive results are in Appendix Tables A3-7*

## 3.1 Effect of fine-tuning and prompt development on model performance, run time and time investment

---

[2] No comprehensive results for sub-10B language models are presented due to their non-significant gain across different settings during the development stage.

[3] This result obtained from Galactica under zero-shot setting; all other results from GPT-JT under few-shot setting.

For the reasoning and classification tasks, fine-tuning BioBERT consistently outperforms the best GPT settings by a considerable margin (ΔF1 0.109 - 0.169) (Fig 2(A) and Table 2). Averaging performance on all datasets, BioBERT's macro F1 scores are significantly better than the other models ($p<0.01$ for all, Fig 2(A). There is no statistical difference between the average performance of the BoW, best performing GPT-3.5s, and GPT-4 models (Appendix Table A3, t-test, see Fig 2a).

Furthermore, engineering the LLM prompts that yield the best results requires a substantial time investment. Fig 2(B) shows the time needed to achieve the best results, taking into account the time needed to develop and identify the prompting strategies for the few-shot settings. Even for the zero-shot setting, which does not require any prompt development, inference time — an indicator of run time and a consideration for compute budget— is longer and yields worse performance compared to BoW and BioBERT performance (Fig 2(B)). Taken together, training a task-specific neural network through classic fine-tuning methodology is both faster and yields better performance.

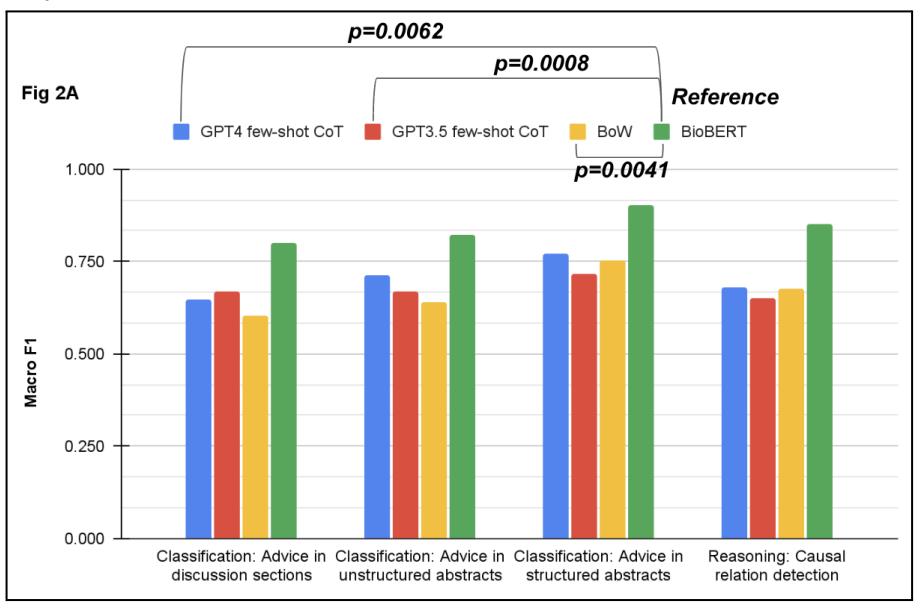

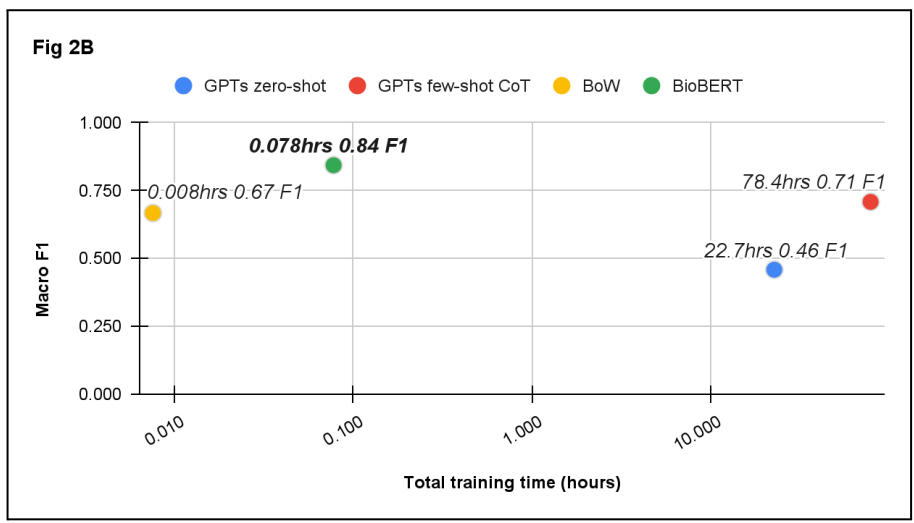

*Figure 2. (A) Comparison of model performance on each dataset. Using fine-tuning BioBERT as a reference, average macro-F1 across all datasets was significantly better than all other models (p<0.05 for all); there was no statistically significant difference in average performance between the other pair-wise model comparisons (Appendix Table A3). (B) Time (hours) required to obtain the best-performing results versus average performance across all datasets. BoW = bag-of-words; CoT = Chain of Thought.*

## 3.2 Effect of amount of training data on model performance

In this experiment, we investigate the amount of training/fine-tuning data from the development set needed to achieve similar or better performing BoW and BioBERT models compared to the best GPT settings (few-shot CoT). As shown in Fig 3, using only 20% of the development set for the supervised fine-tuning of BioBERT surpasses the best GPTs in 3 of 4 datasets. Fine-tuning on 50% of the development set outperforms the best GPTs on all datasets. Furthermore, the simple and computationally efficient BoW model using 100% of the development set outperforms the best GPTs in 2 of 4 datasets.

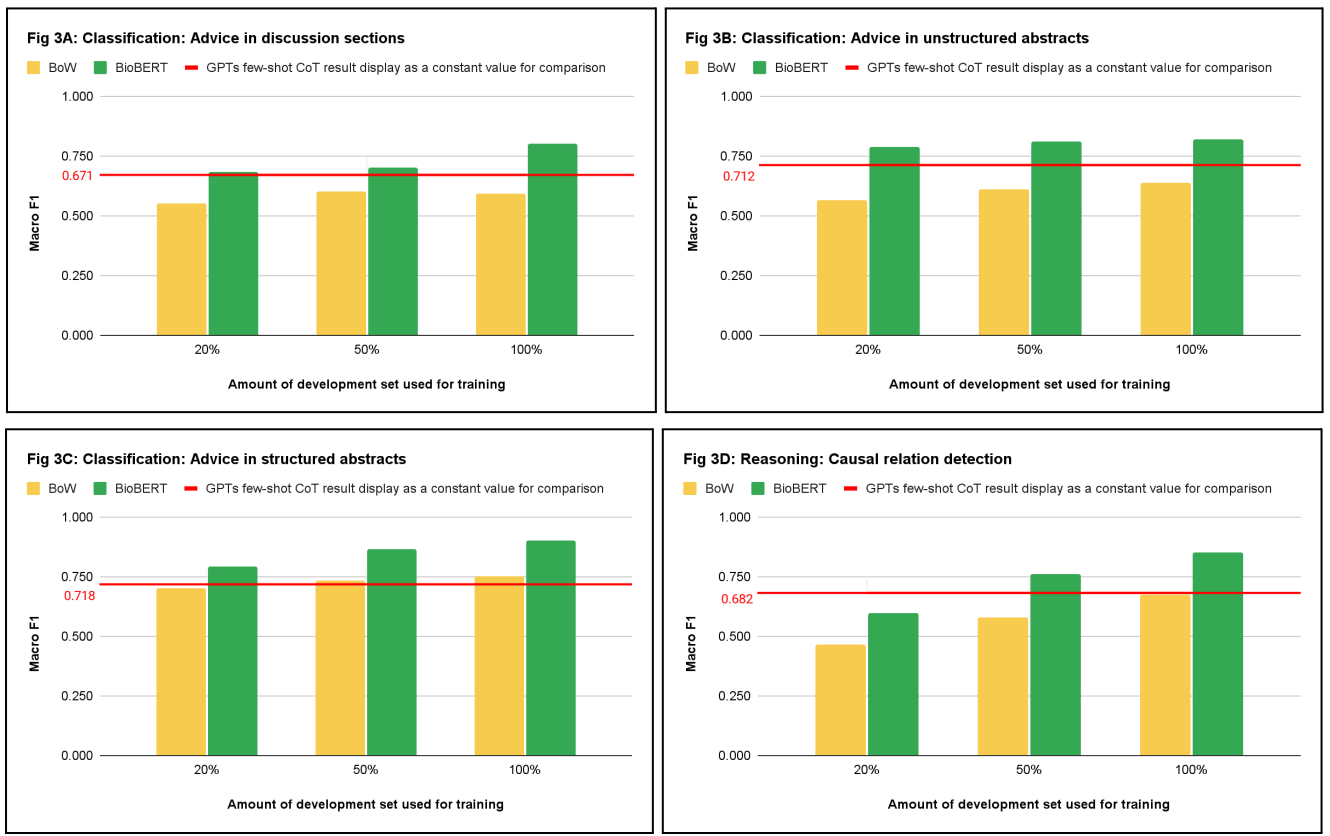

*Figure 3. (A-D) Performance comparison of fine-tuned BioBERT (green bar) and BoW (yellow bar) models with different proportions (20% 50% or 100%) of development set data versus the best GPTs settings (few-shot CoT in red-line) among 4 tasks. (A) Comparison on advice in discussion sections, (B) comparison on advice in unstructured abstracts, (C) comparison on advice in structured abstracts, (D) comparison on causal relation detection. BoW = bag-of-words; CoT = Chain of Thought*

## 3.3 Effect of number of exemplar prompts and CoT prompts on model performance

We examine the relationship between the number of exemplars and CoT prompts and their impact on GPT settings performance. As shown in Fig 4(A), we observe a drop in performance when comparing one-shot to few-shot settings in three of the datasets. Reasoning for causal relation detection is the only task that consistently improves by adding prompt examples, without CoT. However, as demonstrated in Fig 4(B), incorporating more CoT exemplars results in consistent improvements across datasets. This observation highlights the value of adding CoT exemplars, albeit at substantial cost due to the time effort needed to create the CoT prompts.

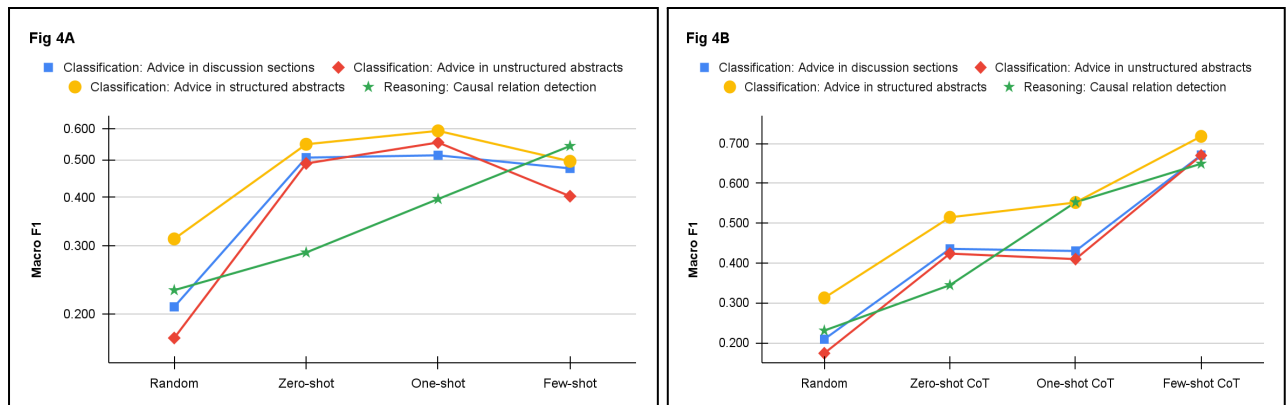

*Figure 4.* Comparison of GPT-3.5s performance on each dataset with increasing exemplars without *(A)* and with *(B)* Chain of Thought (CoT). For both plots, random is shown as a baseline and is the uniform distribution across class labels. Here, one-shot uses the majority class exemplar for each task and few-shot uses one exemplar per class. Few-shot = 3 exemplars for classification datasets and 4 exemplars for the reasoning dataset, reflecting the number of classes for each task.

Another observation from the one-shot experiments (both with and without CoT) among the LLMs evaluated is the impact of different exemplar prompt choices on performance. Variations in the text of the exemplar prompts for one-shot prompts led to notable variations in model outcomes (Appendix Table A6 a-c).

## 4.0 Error Analysis

Of the investigated GPTs, GPT-4 with few-shot CoT settings yields the best performance for 3 of 4 datasets. Thus, we analyze GPT-4's generated CoTs and identify common error patterns to better understand their strengths and limitations. To maintain consistency, we randomly select 100 prediction errors of GPT-4 with few-shot CoT setting from the test set of each dataset. The selection of the error examples follows the same error type ratio from the confusion matrices of the prediction result. Two common error patterns are identified. Pattern A is an incorrect reasoning step based on one specific keyword. For example, the model classifies an input text as a strong advice if the word "importance" appears in the text (Appendix Table A8, row 4). Pattern B is a false positive due to the model incorrectly determining that there is health advice or a relationship for the classification and reasoning tasks, respectively, when in fact there is none. For example, row 3 in Appendix Table A11 presents a pattern B error where GPT-4 misclassifies a relationship between extracted entities as a clinical relationship. Appendix Tables A8-A11 shows examples of the error patterns.

## 5.0 Discussion

In this study, we found that, even with the best in-context learning approaches, fine-tuning BioBERT consistently out-performed LLM performance by macro-F1 >0.100 for all datasets. In fact, fine-tuning on just 20% of the development dataset outperformed the best GPT-4 and GPT-3.5 performance for all of the classification datasets, and outperformed the best GPT-3.5 performance for the reasoning dataset. Surprisingly, the simple BoW models out-performed all LLM in-context learning approaches without CoT, and performed similarly to the best performing GPT-4 approach for classification of structured abstracts (macro-F1 -0.017), and the reasoning task (macro-F1 -0.007).

Our study emphasizes the performance, time, and computational trade-offs that should be taken into account when considering various approaches for clinical NLP tasks. At present, our results suggest that the overall balance is in favor of fine-tuning task-specific smaller models, consistent with Lehman et al[16]. SOTA LLMs such as GPT-3.5/GPT-4 are orders of magnitude larger than traditional language models and cannot be trained or fine-tuned without significant computational resources. For example, GPT-3 has 175 billion parameters and required several thousand petaflop/s-days for pre-training.[24] On the other hand, smaller, more accessible out-of-the-box LLMs without fine-tuning (<10 billion parameters) performed very poorly on our tasks and did not demonstrate improvement with in-context learning, in line with the finding that emergent LLMs abilities scale with model size.[25]

However, the computational requirements of LLMs could theoretically be offset if their zero- or few-shot performance was adequate. Our results clearly demonstrate that zero-short performance was poor. While few-shot CoT prompting improved performance, fine-tuning BioBERT, a 110 million parameter pre-trained language model, consistently provided the best performance. Prompt development to identify the best prompting strategies is itself resource-intensive, requiring human effort to design the prompts, and computational and time resources to evaluate. Ultimately, at most 50% of each dataset's full development set was needed to fine-tune BioBERT models that exceeded LLM performance with the best prompting strategies identified using the full development set. Taking into account prompt development and evaluation, obtaining the best-performing LLM results required 100x the time needed to fine-tune our best performing BioBERT model.

Despite under-performing fine-tuned BioBERT, in-context learning—especially CoT prompting—led to important improvements in performance for the classification and reasoning tasks. The ability of CoT to elicit reasoning and improve LLM performance in the general domain has previously been demonstrated.[22,26] Interestingly, while providing more exemplars with CoT prompting consistently improved performance, this was not always the case for prompting without CoT. For GPT-3.5, one-shot prompting provided the best results for classification, while few-shot prompting provided the best results for reasoning. This could be due to noise provided from including less informative prompts, and highlights the fragility of LLM performance based on the provided prompts.

This lack of robustness is also illustrated by the fact that the choice of exemplar for prompting had major impacts on performance. These findings are in line with Shi et al., who showed that, for arithmetic tasks, prompting may provide irrelevant context that reduces performance.[27] Clearly, simply providing more exemplars does not solve the challenge of LLMs robustness—a major area of concern and future research for the clinical domain, where robust performance is paramount. Concerningly, even small, seemingly non-substantive changes to prompts such as typos have been shown to impact performance.[9,13] There is an emerging body of work on developing strategies to improve robustness and self-consistency.[13,28] Methods to improve performance will be especially important in the medical domain, where jargon, typos, abbreviations, and synonyms are common, limiting our ability to develop reliable prompting strategies and robustly assess real-world performance.

It should be noted that our tasks indirectly address the question of how LLMs perform on clinical NLP text, and use biomedical texts as a proxy for essential clinical NLP tasks. SOTA LLMs such as GPT-3.5/GPT-4 cannot be used with PHI, precluding an evaluation on real clinical data. Nevertheless, LLM performance is known to decrease on out-of-domain tasks, i.e. tasks that include text that does not reflect what it was trained on, including synthetic clinical text.[13] At the same time, classic language models trained on clinical text has been shown to out-perform LLMs with in-context learning.[12] Taken together with our finding that BioBERT, which is pre-trained on biomedical text, out-performs LLMs, it is reasonable to anticipate that findings would be similar on clinical datasets, but further evaluation will be needed once GPT-3.5/GPT-4 are safely and widely accessible for HIPAA-protected data.

Most studies evaluating LLMs for clinical applications have focused on question-answering, with mixed results.[6–9,11] However, question-answering is not representative of the range of NLP tasks needed to process clinical texts,[29] and in isolation has limited practical use for clinic and research. We chose our tasks—classification and reasoning—because they are fundamental to the development of NLP technologies that can support clinical care and research beyond question-answering. NLP for classification entails determining what category an input text belongs in. Classification methods identify if a patient's EMR includes a characteristic or outcome of interest, which has implications for outcomes research, clinical trial matching, and identifying key events at the point-of-care. Reasoning entails determining the relationships between entities, which is to automatically identify how different events in a patient's medical history relate to one another. Especially because nuanced information conveying medical reasoning can often only be expressed in free text, NLP methods for reasoning are needed to automate higher-level medical inferencing. Here, we investigated causative relationships, which is needed for tasks that require linking any clinical outcome with causative factors, such as associating adverse drug events with their inciting agent. Other reasoning tasks include temporal reasoning, which is determining the order of medical events over time. Another benefit of our task selection is that, compared to question-answering, they are more straight-forward to objectively evaluate, enabling a more direct evaluation of how LLMs perform in the high-stakes clinical domain. In the future, evaluation of LLMs in the clinical domain for other classification and reasoning tasks, as well as other common NLP tasks such as relation extraction, named entity recognition, coreference resolution, word sense disambiguation, and machine translation, will be needed.

**6.0 Conclusion**

This study suggests an ongoing role for classic NLP models fine-tuned for specific tasks, while also providing guidance into strategies to optimize the LLMs for the biomedical domain. Fine-tuning BioBERT, a much smaller pretrained language model, out-performed SOTA huge LLMs for biomedical classification and reasoning tasks even after extensive prompt development. CoT prompting with multiple exemplars improved LLM performance compared to zero-shot prompting and prompting without CoT. However, developing CoT prompts was both time- and data-intensive, and BioBERT was more efficient with respect to both measures. In addition, LLMs were very sensitive to prompting strategy and the choice of prompt, raising concerns about the potential to develop LLM methods for medicine that are reliable and safe. Our work provides insight into the potential and pitfalls of these rapidly emerging methods for biomedical text processing. Future research could focus on developing more efficient prompting strategies and fine-tuning techniques for LLMs in the biomedical domain while ensuring their reliability and safety, as well as exploring hybrid approaches that combine the strengths of classic NLP models and LLMs to further enhance performance in biomedical text processing tasks.

**References**

1   Vaswani A, Shazeer N, Parmar N, *et al.* Attention is all you need. In: NeurIPS Proceedings. 2017. https://proceedings.neurips.cc/paper_files/paper/2017/file/3f5ee243547dee91fbd053c1c4a845aa-Paper.pdf (accessed April 4, 2023).

2   Sutton RS, Barto AG. Reinforcement learning: An introduction, 2nd edition. Cambridge, MA: The MIT Press, 2018.

3   Ouyang L, Wu J, Jiang X, *et al.* Training language models to follow instructions with human feedback. In: NeurIPS Proceesings. 2022. https://proceedings.neurips.cc/paper_files/paper/2022/hash/b1efde53be364a73914f58805a001731-Abstract-Conference.html.

4   Lee P, Bubeck S, Petro J. Benefits, Limits, and Risks of GPT-4 as an AI Chatbot for Medicine. *N Engl J Med* 2023; **388**: 1233–9.

5   Reardon S. AI Chatbots Can Diagnose Medical Conditions at Home. How Good Are They? *Scientific American* By Sara Reardon on March 31 2023. https://www.scientificamerican.com/article/ai-chatbots-can-diagnose-medical-conditions-at-home-how-good-are-they/ (accessed April 1, 2023).

6   Gilson A, Safranek CW, Huang T, *et al.* How Does ChatGPT Perform on the United States Medical Licensing Examination? The Implications of Large Language Models for Medical

Education and Knowledge Assessment. *JMIR Med Educ* 2023; **9**: e45312.

7   Liévin V, Hother CE, Winther O. Can large language models reason about medical questions? arXiv [cs.CL]. 2022; published online July 17. http://arxiv.org/abs/2207.08143.

8   Zuccon G, Koopman B. Dr ChatGPT, tell me what I want to hear: How prompt knowledge impacts health answer correctness. arXiv [cs.CL]. 2023; published online Feb 23. http://arxiv.org/abs/2302.13793.

9   Chen S, Kann BH, Foote MB, *et al.* The utility of ChatGPT for cancer treatment information. medRxiv. 2023; : 2023.03.16.23287316.

10  Lyu Q, Tan J, Zapadka ME, *et al.* Translating Radiology Reports into Plain Language using ChatGPT and GPT-4 with Prompt Learning: Promising Results, Limitations, and Potential. arXiv [cs.CL]. 2023; published online March 16. http://arxiv.org/abs/2303.09038.

11  Singhal K, Azizi S, Tu T, *et al.* Large Language Models Encode Clinical Knowledge. arXiv [cs.CL]. 2022; published online Dec 26. http://arxiv.org/abs/2212.13138.

12  Lehman E, Hernandez E, Mahajan D, *et al.* Do We Still Need Clinical Language Models? arXiv [cs.CL]. 2023; published online Feb 16. http://arxiv.org/abs/2302.08091.

13  Wang J, Hu X, Hou W, *et al.* On the Robustness of ChatGPT: An Adversarial and Out-of-distribution Perspective. arXiv [cs.AI]. 2023; published online Feb 22. http://arxiv.org/abs/2302.12095.

14  Liao KP, Cai T, Savova GK, *et al.* Development of phenotype algorithms using electronic medical records and incorporating natural language processing. *BMJ* 2015; **350**: h1885.

15  Zhang Y, Liu M, Zhang L, *et al.* Comparison of Chest Radiograph Captions Based on Natural Language Processing vs Completed by Radiologists. *JAMA Netw Open* 2023; **6**: e2255113.

16  Medori J, Fairon C. Machine learning and features selection for semi-automatic ICD-9-CM encoding. In: Proceedings of the NAACL HLT 2010 Second Louhi Workshop on Text and Data Mining of Health Documents. Los Angeles, California, USA: Association for Computational Linguistics, 2010: 84–9.

17  OpenAI API. http://platform.openai.com (accessed April 3, 2023).

18  Li Y, Wang J, Yu B. Detecting Health Advice in Medical Research Literature. In: Proceedings of the 2021 Conference on Empirical Methods in Natural Language Processing. Online and Punta Cana, Dominican Republic: Association for Computational Linguistics, 2021: 6018–29.

19  Yu B, Li Y, Wang J. Detecting Causal Language Use in Science Findings. In: Proceedings of the 2019 Conference on Empirical Methods in Natural Language Processing and the 9th International Joint Conference on Natural Language Processing (EMNLP-IJCNLP). Hong Kong, China: Association for Computational Linguistics, 2019: 4664–74.

20  Devlin J, Chang M-W, Lee K, Toutanova K. BERT: Pre-training of Deep Bidirectional Transformers for Language Understanding. arXiv [cs.CL]. 2018; published online Oct 11. http://arxiv.org/abs/1810.04805.

21  Lee J, Yoon W, Kim S, *et al.* BioBERT: a pre-trained biomedical language representation model for biomedical text mining. *Bioinformatics* 2020; **36**: 1234–40.

22  Wei J, Wang X, Schuurmans D, *et al.* Chain-of-Thought Prompting Elicits Reasoning in Large Language Models. arXiv [cs.CL]. 2022; published online Jan 28. http://arxiv.org/abs/2201.11903.

23  Taylor R, Kardas M, Cucurull G, *et al.* Galactica: A Large Language Model for Science. arXiv [cs.CL]. 2022; published online Nov 16. http://arxiv.org/abs/2211.09085.

24  Brown TB, Mann B, Ryder N, *et al.* Language Models are Few-Shot Learners. arXiv [cs.CL]. 2020; published online May 28. http://arxiv.org/abs/2005.14165.

25  Wei J, Tay Y, Bommasani R, *et al.* Emergent Abilities of Large Language Models. arXiv [cs.CL]. 2022; published online June 15. http://arxiv.org/abs/2206.07682.

26  Kojima T, Gu SS, Reid M, Matsuo Y, Iwasawa Y. Large Language Models are Zero-Shot Reasoners. arXiv [cs.CL]. 2022; published online May 24. http://arxiv.org/abs/2205.11916.

27  Shi F, Chen X, Misra K, *et al.* Large Language Models Can Be Easily Distracted by Irrelevant Context. arXiv [cs.CL]. 2023; published online Jan 31. http://arxiv.org/abs/2302.00093.

28  Wang X, Wei J, Schuurmans D, *et al.* Self-Consistency Improves Chain of Thought Reasoning in Language Models. arXiv [cs.CL]. 2022; published online March 21. http://arxiv.org/abs/2203.11171.

29  Savova GK, Danciu I, Alamudun F, *et al.* Use of Natural Language Processing to Extract Clinical Cancer Phenotypes from Electronic Medical Records. *Cancer Res* 2019; **79**: 5463–70.

**Appendix A:**

|  | Classification: Advice in discussion sections | Classification: Advice in unstructured abstracts | Classification: Advice in structured abstracts | Reasoning: Causal relation detection |
|---|---|---|---|---|
| random seed | 42 | 42 | 42 | 42 |
| batch size (per_device_train) | 128 | 128 | 128 | 128 |
| batch size (per_device_eval) | 512 | 512 | 512 | 512 |
| epoch | 50 | 50 | 50 | 200 |
| learning rate | 2e-5 | 2e-5 | 2e-5 | 2e-5 |
| max sequence length | 512 | 512 | 512 | 512 |
| warmup_steps | 500 | 500 | 500 | 500 |
| weight_decay | 0.01 | 0.01 | 0.01 | 0.01 |

*Table A1. Parameters and hyperparameters for fine-tuning BioBERT*

Tables A2 a-c provide a comparison of the estimated time required for training and inferencing using different models and settings.
The total spending of inference OpenAI API was $1,299 based on OpenAI pricing for Feb-Mar, 2023[4]

| Model & Setting | Prompt creation time (in seconds) | Inference time for training on dev set (in seconds) | Inference time on test set (in seconds) | Total (in hours) | Performance |
|---|---|---|---|---|---|
| GPT zero-shot | 2400.000 | 28296.000 | 11322.000 | 11.005 | 0.458 |
| GPT few-shot CoT | 18000.000 | 94320.000 | 37740.000 | 36.683 | 0.709 |
| BoW classifier | 0s | 13.000 | 0.700 | 0.004 | 0.667 |
| BioBERT FT | 0s | 139.427 | 0.800 | 0.039 | 0.844 |
|  | It took roughly about 240s for creating zero-shot prompts. And it took roughly 1800s for a few-shot CoT prompt writing time. And there were a total of 10 prompts to test. | | | | |
|  | Inference time is 1.2s per query for 0-shot, 4s per query 4-shot CoT. | | | | |
|  | Performance is calculated by average Macro F1 among the four tasks given setting. | | | | |

*Table A2a. Time required for training and inferencing of each method*

---

[4] https://openai.com/pricing

| Model & Setting | Oracle prompter | Assuming an Oracle prompt writer who creates only one prompt to achieve the best performance on the test set. Thus, no prompts needed for the validation set. | | | |
|---|---|---|---|---|---|
| | Prompt creation time (in seconds) | Inference time for training on dev set (in seconds) | Inference time on test set (in seconds) | Total (in hours) | Performance |
| GPT zero-shot | 120.000 | 0.000 | 11322.000 | 3.145 | 0.458 |
| GPT few-shot CoT | 900.000 | 0.000 | 37740.000 | 10.483 | 0.709 |
| BoW classifier | 0.000 | 13.000 | 0.700 | 0.004 | 0.667 |
| BioBERT FT | 0.000 | 139.427 | 0.800 | 0.039 | 0.844 |
| | Note that: we assume such Oracle prompt writer will also create prompts 100% more efficient than us. | | | | |

*Table A2b. Time required for training and inferencing of an Oracle prompter*

| Model & Setting | Oracle prompter | Batch mode: 512 | | | |
|---|---|---|---|---|---|
| | Prompt creation time (in seconds) | Inference time for training on dev set (in seconds) | Inference time on test set (in seconds) | Total (in hours) | Performance |
| GPT zero-shot | 120.000 | 0.000 | 22.113 | 0.006 | 0.458 |
| GPT few-shot CoT | 900.000 | 0.000 | 73.711 | 0.020 | 0.709 |
| BoW classifier | 0.000 | 13.000 | 0.700 | 0.004 | 0.667 |
| BioBERT FT | 0.000 | 139.427 | 0.800 | 0.039 | 0.844 |

*Table A2c. Time required for training and inferencing of an Oracle prompter with machine that can parallel inference in batch of 512*

| | GPT-3.5s | BoW | BioBERT |
|---|---|---|---|
| GPT4 | 0.4150 | 0.4228 | 0.0062** |
| GPT-3.5s | – | 0.7969 | 0.0008** |
| BoW | – | – | 0.0041** |

**: $p<0.05$

*Table A3. P-values for models significance comparisons (t-test)*

|  | Classification: Advice in discussion sections | Classification: Advice in unstructured abstracts | Classification: Advice in structured abstracts | Reasoning: Causal relation detection |
|---|---|---|---|---|
| Zero-shot Davinci | 0.503 | 0.461 | **0.548** | 0.242 |
| Zero-shot Turbo | 0.506 | **0.489** | 0.484 | 0.288 |
| Zero-shot GPT4 | **0.509** | 0.478 | 0.517 | **0.304** |
| Few-shot CoT Davinci | **0.671** | 0.617 | 0.718 | 0.649 |
| Few-shot CoT Turbo | 0.603 | 0.670 | 0.699 | 0.566 |
| Few-shot CoT GPT4 | 0.648 | **0.712** | **0.770** | **0.682** |

*Table A4. GPT3.5-Davinci vs Turbo(ChatGPT) vs GPT4 comparison results on the test set. Best results are **bolded**.*

|  | Classification: Advice in discussion sections | Classification: Advice in unstructured abstracts | Classification: Advice in structured abstracts | Reasoning: Causal relation detection |
|---|---|---|---|---|
| Zero-shot Davinci | 0.503 | 0.461 | 0.548 | 0.242 |
| Zero-shot Turbo | **0.506** | **0.489** | 0.484 | **0.288** |
| One-shot Davinci | 0.513 | 0.501 | 0.593 | 0.210 |
| One-shot Turbo | 0.513 | **0.554** | 0.470 | **0.396** |
| Few-shot Davinci | 0.468 | 0.403 | 0.495 | 0.542 |
| Few-shot Turbo | **0.475** | 0.243 | 0.436 | 0.513 |
| Zero-CoT Davinci | 0.436 | 0.383 | 0.515 | 0.332 |
| Zero-CoT Turbo | 0.418 | **0.424** | 0.486 | **0.345** |
| One-CoT Davinci | 0.418 | 0.423 | 0.470 | 0.489 |
| One-CoT Turbo | 0.430 | 0.410 | 0.552 | 0.553 |
| Few-CoT Davinci | 0.671 | 0.617 | 0.718 | 0.649 |
| Few-CoT Turbo | **0.603** | **0.670** | 0.699 | **0.566** |

*Table A5. GPT3.5-Davinci vs Turbo(ChatGPT) comparison results on the test set. **Bolded** indicates Turbo performs better than Davinci (10 out of 24 settings), notably consistent better on Classification: Advice in unstructured abstracts.*

Tables A6 (a-c) present a comparison of development/test set performance when exemplars of class choice in one-shot CoT prompts are varied using GPT3.5-Davinci. The systematic analysis of distinct class exemplars highlights the effectiveness of various exemplar selections in the CoT approach. Table A6b-c shows the change in performance on the development/test set when a different exemplar is used.

| Class | Classification: Advice in discussion sections | Classification: Advice in unstructured abstracts | Classification: Advice in structured abstracts | Reasoning: Causal relation detection |
|---|---|---|---|---|
| None class/**Majority** | 0.421 | **0.410** | **0.552** | 0.526 |
| Weak/Correlational | **0.430** | 0.283 | 0.381 | 0.455 |
| Strong/Conditional Causal | 0.343 | 0.319 | 0.447 | **0.553** |
| Direct Causal | – | – | – | 0.386 |

*Table A6a. GPT3.5-Davinci, one-shot CoT results on the test set.*
***Bold** indicates best performance.*

| Class | Classification: Advice in discussion sections | Classification: Advice in unstructured abstracts | Classification: Advice in structured abstracts | Reasoning: Causal relation detection |
|---|---|---|---|---|
| None class/**Majority** | **0.462** | **0.510** | **0.552** | **0.521** |
| Weak/Correlational | 0.460 | 0.333 | 0.391 | 0.495 |
| Strong/Conditional Causal | 0.393 | 0.339 | 0.448 | 0.453 |
| Direct Causal | – | – | – | 0.456 |

*Table A6b. GPT3.5-Davinci, one-shot results on the development set.*
***Bold** indicates best performance.*

| Class | Classification: Advice in discussion sections | Classification: Advice in unstructured abstracts | Classification: Advice in structured abstracts | Reasoning: Causal relation detection |
|---|---|---|---|---|
| None class/**Majority** | **0.411** | 0.390 | **0.423** | 0.526 |
| Weak/Correlational | 0.340 | 0.283 | 0.361 | 0.455 |
| Strong/Conditional Causal | 0.343 | **0.430** | 0.350 | **0.553** |
| Direct Causal | – | – | – | 0.456 |

*Table A6c. GPT3.5-Davinci, one-shot CoT results on the development set.*
***Bold** indicates best performance.*

|                      | Classification: Advice in discussion sections | Classification: Advice in unstructured abstracts | Classification: Advice in structured abstracts | Reasoning: Causal relation detection |
|---|---|---|---|---|
| Zero-shot GPT-J      | 0.248 | 0.230 | 0.302 | 0.211 |
| Zero-shot GPT-JT     | **0.272** | **0.262** | **0.321** | 0.226 |
| Zero-shot Galactica  | 0.234 | 0.218 | 0.294 | **0.252** |
| Few-shot GPT-J       | 0.157 | 0.142 | 0.292 | 0.191 |
| Few-shot GPT-JT      | **0.276** | **0.284** | **0.358** | 0.216 |
| Few-shot Galactica   | 0.246 | 0.205 | 0.293 | **0.244** |

*Table A7a. Performance of smaller models (less than 10B parameters) on the test set. Results with the best two settings shown in experiments on the dev set. Best results are **bolded**.*

|                      | Classification: Advice in discussion sections | Classification: Advice in unstructured abstracts | Classification: Advice in structured abstracts | Reasoning: Causal relation detection |
|---|---|---|---|---|
| Zero-shot GPT-J      | 0.247 | 0.246 | 0.296 | 0.210 |
| Zero-shot GPT-JT     | **0.285** | **0.262** | **0.331** | 0.226 |
| Zero-shot Galactica  | 0.233 | 0.209 | 0.278 | **0.250** |
| Few-shot GPT-J       | 0.155 | 0.143 | 0.286 | 0.192 |
| Few-shot GPT-JT      | **0.284** | **0.282** | **0.348** | 0.217 |
| Few-shot Galactica   | 0.257 | 0.222 | 0.308 | **0.257** |
| Few-CoT GPT-J        | 0.211 | 0.200 | 0.323 | – |
| Few-CoT GPT-JT       | 0.172 | 0.129 | 0.269 | – |
| Few-CoT Galactica    | 0.241 | 0.195 | 0.368 | – |

*Table A7b. Performance of smaller models (less than 10B parameters) on the dev set. Best results are **bolded**.*

| Sentence | GPT–4 Answer | Gold | Pred |
|---|---|---|---|
| Testing this hypothesis may represent a direction of future research. | 1. The term "may represent" indicates that there is an advice in the context. 2. "May represent" suggests a possibility, which is not a strong opinion. 3. The answer is weak advice. | no advice | weak advice |
| Future research should attempt to replicate these findings in a larger-scale trial using a more sophisticated, user-friendly program and additional measures of improvement in more diverse samples. | 1. The term "should attempt" indicates that there is an advice in the context. 2. "Should" indicates a directive opinion. 3. The answer is strong advice. | no advice | strong advice |
| Our results suggest that pregnant women with an HbA1c of 5.4% (36 mmol/mol) should proceed with an OGTT. | 1. The term "suggest" indicates that there is an advice in the context. 2. "Suggest" indicates a directive opinion, but not as strong as a recommendation. 3. The answer is weak advice. | strong advice | weak advice |
| This highlights the importance of considering subgroups of patients with LRTI to develop RP evidence-based practice. | 1. The context mentions the importance of considering subgroups, which implies a suggestion or advice. 2. The term "importance" indicates that the opinion is not weak. 3. The answer is strong advice. | weak advice | strong advice |
| Surgery done as soon as possible is helpful to reduce the force caused by the obstruction and helpful to the normal facial growth. | 1. The context states that surgery is helpful, which implies a suggestion or advice. 2. The phrase "as soon as possible" indicates a sense of urgency, making the advice stronger. 3. The answer is strong advice. | weak advice | strong advice |
| This study shows the lack of normative consensus about ACP and highlights the need for consistent educational programs regarding the role of the CBLTC provider in the ACP process. | 1. The context states the need for consistent educational programs, which is an advice. 2. The term "highlights the need" indicates that the opinion is not strong. 3. The answer is weak advice. | strong advice | weak advice |
| Pattern A | Incorrect reasoning step based on specific keywords | | |
| Pattern B | Incorrect classification of health advice | | |

Table A8: Sample of generated three-shot Chain-of-Thought from GPT-4 for advice in the structured abstracts in the HealthAdvice test set.

| Sentence | GPT-4 Answer | Gold | Pred |
|---|---|---|---|
| Future research should seek to implement additional evidence-based actions to improve QoL in this population. | 1. The term "should seek to implement" indicates that there is an advice in the context. 2. "Should" indicates a strong opinion. 3. The answer is strong advice. | no advice | strong advice |
| In conclusion, contrast-enhanced whole-heart CMRA at 3.0T can depict the normal and variant cardiac venous anatomy. | 1. The context states a conclusion, which is not an advice. 2. The answer is no advice. | weak advice | no advice |
| However, because our analysis was based on a relatively small number of genes, a larger study will be needed to identify the best predictors of disease outcome overall. | 1. The context mentions "a larger study will be needed," which indicates an advice. 2. "Will be needed" suggests a strong opinion. 3. The answer is strong advice. | no advice | strong advice |
| The manufacturer of the CDI 500 in-line blood gas monitor recommends gas calibration before use. | 1. The context states a recommendation, which is a form of advice. 2. The term "recommends" indicates a strong opinion. 3. The answer is strong advice. | no advice | strong advice |
| Health promotion in transport industry workplaces should be an important topic for future research. | 1. The term "should be" indicates that there is an advice in the context. 2. "Should be" indicates a strong opinion. 3. The answer is strong advice. | no advice | strong advice |
| Treatment with iodoform would be beneficial. | 1. The term "would be beneficial" indicates an advice in the context. 2. "Would be" indicates a strong opinion. 3. The answer is strong advice. | weak advice | strong advice |
| A comprehensive early childhood vision screening program in Hawai'i utilizing automated hand-held | 1. The term "may have" indicates an advice in the context. 2. "May have" indicates a weak opinion. | no advice | weak advice |

| Sentence | GPT-4 Answer | | |
|---|---|---|---|
| photoscreeners may have public health value. | 3. The answer is weak advice. | | |
| Although the AD-associated genetic variants appear to have small effects on early cognitive change, replication will be required to establish whether there is a discernible influence on cognitive status in midlife. | 1. The phrase "replication will be required" indicates an advice in the context.<br>2. "Will be required" indicates a strong opinion.<br>3. The answer is strong advice. | no advice | strong advice |
| Replication tests in more generalizable samples and of longer duration are indicated. | 1. The term "are indicated" suggests that there is an advice in the context.<br>2. "Are indicated" implies a strong opinion.<br>3. The answer is strong advice. | no advice | strong advice |
| Pattern A | Incorrect reasoning step based on specific keywords | | |
| Pattern B | Incorrect classification of health advice | | |

*Table A9: Sample of generated three-shot Chain-of-Thought from GPT-4 for advice in the unstructured abstracts in the HealthAdvice test set.*

| Sentence | GPT-4 Answer | Gold | Pred |
|---|---|---|---|
| In this Saudi sample, the mean arch widths were narrower when compared to a Turkish sample,[19] a North American sample, a South American sample, a Korean sample,[20] and an Egyptian sample;[21] on the other hand, it was close to a Malay sample.[22] These ethnic groups' differences in arch dimensions explain the need for specific orthodontic arch wire for each patient, based on the initial arch form. | 1) The claim is discussing the differences in arch dimensions among various ethnic groups.<br>2) The claim suggests the need for specific orthodontic arch wires for each patient based on their initial arch form.<br>3) The claim does not use strong or hedged language, but it does imply an actionable suggestion for orthodontic treatment.<br>4) Therefore, the claim has weak advice.<br>Answer: (2) weak advice. | no advice | weak advice |
| In comparison, Patient 3 demonstrated significant PR3-ANCA antibody production without evidence of peripheral B cell repopulation, which emphasizes the need to consider granulomata in GPA as B cell containing tertiary lymphoid structures pivotal to driving the relapsing disease course. | 1) The claim is discussing the importance of considering granulomata in GPA as B cell containing tertiary lymphoid structures.<br>2) The claim uses the phrase "emphasizes the need to consider," which indicates a suggestion or recommendation.<br>3) However, the strength of the advice is not explicitly strong or weak, but it leans more towards a weak advice due to the use of the word "emphasizes."<br>Answer: (2) weak advice. | strong advice | weak advice |
| Despite considerable challenges that face the interpretation of clinicoanatomical correlations in neurodegenerative diseases, the current results suggest that at least parts of the left anterior temporal neocortex should be inserted into the temporosylvian language network where they are likely to play critical roles in linking words to their object referents and in sharpening the taxonomic acuity of naming and word comprehension. | 1) The claim has an actionable suggestion related to the insertion of parts of the left anterior temporal neocortex into the temporosylvian language network.<br>2) When describing the suggestion, the authors used the phrase "suggest that" and "likely to play critical roles", which indicates the strength of the claim is not strong but weak.<br>3) Therefore, the claim has weak advice.<br>Answer: (2) weak advice. | strong advice | weak advice |
| The results, coupled with other correlational and quasiexperimental results, stress the importance of providing high quality evidence-based reading instruction throughout the early grades. | 1) The claim has an actionable suggestion related to providing high-quality, evidence-based reading instruction in early grades.<br>2) When describing the suggestion, the authors used the phrase "stress the importance of", which indicates the strength of the claim is strong.<br>3) Therefore, the claim has strong advice.<br>Answer: (1) strong advice. | weak advice | strong advice |
| It is important for nurses to be aware of rescue analgesia strategies, such as PCA analgesia to protect against breakthrough pain on ambulation, end-of-dose failure on controlled-release opioids, or spontaneous pain with no obvious pathology (Chiaretti & Langer, 2005). | 1) The claim has an actionable suggestion for nurses to be aware of rescue analgesia strategies.<br>2) When describing the suggestion, the authors used the phrase "it is important for", which indicates the strength of the suggestion is strong.<br>3) Therefore, the claim has strong advice.<br>Answer: (1) strong advice. | weak advice | strong advice |

| Sentence | GPT-4 Answer | | |
|---|---|---|---|
| It is reasonable for IVF clinics to use low oxygen concentration from embryo culture beginning, and it will optimize the patients' treatment outcome in a long-term perspective. | 1) The claim has an actionable suggestion related to the use of low oxygen concentration in IVF clinics.<br>2) When describing the suggestion, the authors used the word "reasonable", which indicates a certain level of confidence in the suggestion, making it strong advice.<br>3) Therefore, the claim has strong advice.<br>Answer: (1) strong advice. | weak advice | strong advice |
| While the study findings provide information about QoL and some influencing factors among PLWHIV in Georgia, the limitations of the current study can be considered and minimized in future studies. | 1) The claim is talking about the study findings and their limitations.<br>2) The claim suggests that the limitations can be considered and minimized in future studies, which is an actionable suggestion for future research.<br>3) However, the claim does not use strong modal words or phrases, making the advice weak.<br>Answer: (2) weak advice. | no advice | weak advice |
| Pattern A    Incorrect reasoning step based on specific keywords<br>Pattern B    Incorrect classification of health advice | | | |

*Table A10: Sample of generated three-shot Chain-of-Thought from GPT-4 for advice in the discussion in the HealthAdvice test set.*

| Sentence | GPT-4 Answer | Gold | Pred |
|---|---|---|---|
| The present literature supports the idea that authoritative parenting may be protective against later overweight and obesity, although findings are mixed. | 1) The sentence describes the relationship between "authoritative parenting" and "later overweight and obesity".<br>2) The authors used the language cue "protective against", which suggests a relationship between the two entities.<br>3) However, the authors also mentioned that "findings are mixed", which indicates that the relationship is not definitive or certain.<br>4) Therefore, the sentence expresses a correlation, but with some uncertainty.<br>Answer: 2 - correlation | conditional causation | correlation |
| The bidirectional association between AN diagnosis and CD warrants attention in the initial assessment and follow-up of these conditions because underdiagnosis and misdiagnosis of these disorders likely cause protracted and unnecessary morbidity. | 1) The sentence describes the relationship between "AN diagnosis" and "CD".<br>2) The authors used the phrase "bidirectional association", which suggests a correlation between the two conditions.<br>3) Therefore, the sentence expresses correlation.<br>Answer: 2 - correlation | conditional causation | correlation |
| This figure would be below accepted international thresholds to undertake screening mammography and raises radiation protection issues. | 1) The sentence describes the relationship between "the figure being below accepted international thresholds" and "raising radiation protection issues".<br>2) The authors used the verb "raises", which suggests that there is a direct effect between them.<br>3) When describing the effect, the authors did not use any hedge words or modal verbs to tone down the certainty of the effect.<br>4) Therefore, the sentence expresses direct causation.<br>Answer: 4 - causation | no relation | causation |
| High-risk populations may accept CESM as a screening exam and may prefer it over screening MRI if ongoing trials demonstrate screening CESM to be clinically non-inferior MRI. | 1) The sentence describes the relationship between "accepting CESM as a screening exam" and "ongoing trials demonstrating screening CESM to be clinically non-inferior MRI".<br>2) The authors used the conditional language "may accept" and "if ongoing trials demonstrate", which suggests that there is a conditional effect between them.<br>3) Therefore, the sentence expresses conditional causation.<br>Answer: 3 - conditional causation | no relation | conditional causation |
| 3D MRI reconstruction after NAC could accurately detect the residual tumors after neoadjuvant chemotherapy, and contribute to select patients who received breast conserving therapy after NAC with tumor downstaging. | 1) The sentence describes the relationship between "3D MRI reconstruction after NAC" and "accurately detecting residual tumors" and "contributing to selecting patients for breast conserving therapy after NAC with tumor downstaging".<br>2) The authors used the verbs "could accurately detect" and "contribute to select", which suggest that there is a direct effect between them.<br>3) When describing the effect, the authors also used a hedge word "could" to tone down the certainty of the effect.<br>4) Therefore, the sentence expresses conditional causation only.<br>Answer: 3 - conditional causation | direct causation | conditional causation |

| Sentence | Reasoning | Prediction | Gold |
|---|---|---|---|
| The results showed no effects of aromatherapy on QOL, sleep quality, and vital signs. | 1) The sentence describes the relationship between "aromatherapy" and "QOL, sleep quality, and vital sign". <br> 2) The authors used the phrase "no effects", which suggests that there is no relationship between the two entities. <br> 3) Therefore, the sentence expresses no relationship. <br> Answer: 1 - no relationship | direct causation | no relation |
| The rs7903146 (C/T) polymorphism of the TCF7L2 gene might not be associated with obesity in the Cameroonian population. | 1) The sentence describes the relationship between "rs7903146 (C/T) polymorphism of the TCF7L2 gene" and "obesity in the Cameroonian population". <br> 2) The authors used the language cue "might not be associated with" to describe the relationship, which suggests a lack of correlation. <br> 3) The use of "might not" indicates uncertainty, but the focus is still on the potential absence of a relationship. <br> 4) Therefore, the sentence expresses no relationship. <br> Answer: 1 - no relationship | correlation | no relation |
| This protocol will save money, time, and workload without compromising patients safety and future management. | 1) The sentence describes the relationship between "this protocol" and "saving money, time, and workload" as well as "not compromising patient safety and future management". <br> 2) The authors used the verb "will save" and "without compromising", which suggests that there is a direct effect between the protocol and the outcomes mentioned. <br> 3) The certainty of the effect is strong and direct, as the authors used the word "will" to describe the effect. <br> 4) Therefore, the sentence expresses causation. <br> Answer: 4 - causation | no relation | causation |
| We should have major policy changes to increase future screening programs and health education programs which would have an overall positive impact on reducing the disease burden. | 1) The sentence describes the relationship between "major policy changes" and "reducing the disease burden". <br> 2) The authors used the language cue "would have an overall positive impact on", which suggests that there is a direct effect between them. <br> 3) When describing the effect, the authors used a hedge word "would" to tone down the certainty of the effect. <br> 4) Therefore, the sentence expresses conditional causation only. <br> Answer: 3 - conditional causation | no relation | conditional causation |
| Pattern A | Incorrect reasoning step based on specific keywords | | |
| Pattern B | Incorrect classification of reasoning relation | | |

Table A11: Sample of generated four-shot Chain-of-Thought from GPT-4 for reasoning in the CausalRelation test set.

# Appendix B:

| Prompt Template | Used for | ID | Dataset |
|---|---|---|---|
| "Context: {context} \n Question: is this a 2) strong advice, 1) weak advice 0) no advice?" | Zero-shot | cs | HealthAdvice |
| "Context: {context} \n Question: does the context contain any medical advice? : 2: yes, 1: maybe 0: no" | | cs | HealthAdvice |
| Context: {context} \n Question: Is it a (0) no advice, (1) strong advice, or (2) weak advice statement? | | yy | HealthAdvice |
| Context: {context} \n Question: Does this claim have (1) strong advice, (2) weak advice, or there is (0) no advice? | One-shot | yy | HealthAdvice |
| Context: {context} \n Question: Is this (0) no advice, (1) strong advice, or (2) weak advice? | Few-shot | db | HealthAdvice |
| Context: {context} \n Question: What type of advice is this? Select only one from: 0 – no advice, 1 - strong advice, or 2 - weak advice. | | db | HealthAdvice |
| Context: {context} \n Label the sentence as strong medical advice, weak medical advice or no medical advice | Zero-shot-CoT | gs | HealthAdvice |
| Context: {context} \n The strength of the medical advice if any in this sentence is .... | | gs | HealthAdvice |
| Context: {context} -- Question: What type of relationship is this describing? Select only one from: 1) no relationship, 2) correlation, 3) conditional causation, or 4) direct causation. | | hv | HealthAdvice |
| Context: {context} -- Question: Is the previous statement describing a (1) directly correlative relationship, (2) conditionally causative relationship, (3) causative relationship, or (0) no relationship. | | hv | HealthAdvice |
| Context: {context} \n Question: choose from the following causal relationships: 0: None, 1: Correlational, 2: Conditional causal, 3: Direct causal? | | cs | CausalRelation |
| Context: {context} \n Question: Is this a: 0) None, 1) Correlational, 2) Conditional causal, 3) Direct causal?" | | cs | CausalRelation |
| Context: {context} \n Question: Does 1 - correlation, 2 - conditional causation, or 3 – direct causation expressed in the sentence, or it is a 0 - no relationship sentence? | Few-shot | yy | CausalRelation |
| Context: {context} \n Question: Is it 1 - no relationship, 2 - correlation, 3 - conditional causation, or 4 – direct causation? | One-shot | yy | CausalRelation |
| Context: {context} \n Question: What type of relationship is this describing? Select only one from: 0 - no relationship, 2 - correlation, 3 - conditional causation, or 4 – direct causation. | Zero-shot Zero-shot-CoT | db | CausalRelation |
| Context: {context} \n Question: Is this describing a (1) directly correlative relationship, (2) conditionally causative relationship, (3) causative relationship, or (0) no relationship. | | db | CausalRelation |
| Context: {context} \n Label the relation expressed in the sentence as one of correlation, conditional causation, causation or other | | gs | CausalRelation |
| Context: {context} \n The relation in the sentence is of type... | | gs | CausalRelation |
| Context: {context} -- Question: What type of relationship is this describing? Select only one from: 1) no relationship, 2) correlation, 3) conditional causation, or 4) direct causation. | | hv | CausalRelation |
| Context: {context} -- Question: Is the previous statement describing a (1) directly correlative relationship, (2) conditionally causative relationship, (3) causative relationship, or (0) no relationship. | | hv | CausalRelation |

*Table B1: Prompt templates, no CoT. Second column indicates what prompt was selected for each dataset.*

| Prompt Template | Dataset | Setting |
|---|---|---|
| "CONTEXT: Third, although our study have taken a large number of potential confounders into consideration, we could not completely rule out the possibility of unmeasured confounding.<br>QUESTION: Does this claim have (1) strong advice, (2) weak advice, or there is (0) no advice?<br>Let's think step by step:<br>1) The claim is talking about limitations of the current study.<br>2) The claim does not have any actionable suggestions for health-related clinical or policy changes.<br>3) Therefore, the claim does not have strong or weak advice. It is a no advice statement.<br>Answer: (0) no advice.<br><br>CONTEXT: Georgian public health specialists working in the HIV field should prioritize implementation of such interventions among HIV patients.<br>QUESTION: Does this claim have (1) strong advice, (2) weak advice, or there is (0) no advice?<br>Let's think step by step:<br>1) The claim has an actionable suggestion related to the implementation of an intervention.<br>2) When describing the suggestion, the authors used modal word "should", which indicates the strength of the suggestion and it is strong.<br>3) Therefore, the claim has strong advice.<br>Answer: (1) strong advice.<br><br>CONTEXT: Here we demonstrate that cancer recurrence after curative surgery was significantly lower in ANP-treated patients than in control patients, suggesting that ANP could potentially be used to prevent cancer recurrence after surgery.<br>QUESTION: Does this claim have (1) strong advice, (2) weak advice, or there is (0) no advice?<br>Let's think step by step:<br>1) The claim has an actionable suggestion for the use of ANP to prevent cancer recurrence after surgery.<br>2) When describing the suggestion, the authors used a hedged phrase "could potentially be used to", which indicates the strength of the claim is not strong but weak.<br>3) Therefore, the claim has weak advice.<br>Answer: (2) weak advice.<br><br>CONTEXT: {context}<br>QUESTION: Does this claim have (1) strong advice, (2) weak advice, or there is (0) no advice?<br>Let's think step by step: | Classification: Advice in discussion sections | Few-shot CoT |
| CONTEXT: Correlation of serologic titers for Chlamydia trachomatis with other tests has been based on direct fluorescence antibody (DFA) testing and culture, but not on nucleic acid-based tests that are used for screening.<br>QUESTION: Is this a (0) no advice, (1) weak advice, or (2) strong advice? Let's think step by step.<br>ANSWER:<br>1. The context states a fact, which is not an advice.<br>2. The answer is no advice.<br><br>CONTEXT: These results suggest that the LOS test is an informative tool that should be included in any objective balance evaluations that screen TBI patients with balance complaints.<br>QUESTION: Is this a (0) no advice, (1) weak advice, or (2) strong advice? Let's think step by step.<br>ANSWER:<br>1. The term "should be included" indicate that there is an advice in the context.<br>2. "Should be" indicates a strong opinion.<br>3. The answer is strong advice.<br><br>CONTEXT: Therefore, intraoperative antifibrinolysis may not be indicated in routine cardiac surgery when other blood-saving techniques are adopted.<br>QUESTION: Is this a (0) no advice, (1) weak advice, or (2) strong advice? Let's think step by step.<br>ANSWER:<br>1. The term "may not be indicated" indicates an advice in the context.<br>2. "May not" indicates a weak opinion.<br>3. The answer is weak advice.<br><br>CONTEXT: {context}<br>QUESTION: Is this a (0) no advice, (1) weak advice, or (2) strong advice? Let's think step by step. | Classification: Advice in unstructured abstracts | Few-shot CoT |
| CONTEXT: Further mechanistic research in larger cohorts is necessary to reconcile the potential role of T2D in UF risk.<br>QUESTION: Is this a (0) no advice, (1) weak advice, or (2) strong advice? Let's think step by step.<br>ANSWER:<br>1. The context states the necessity of further research, which is not an advice.<br>2. The answer is no advice.<br><br>CONTEXT: Since blood pressure problems run a worse course in Blacks, we recommend encouragement of night-time intake in those preferring it and suggest that in those requiring two or more drugs one should be taken at night.<br>QUESTION: Is this a (0) no advice, (1) weak advice, or (2) strong advice? Let's think step by step.<br>ANSWER:<br>1. The terms "we recommend" and "suggest" indicate that there is an advice in the context.<br>2. "Recommend" and "suggest" indicate a directive opinion.<br>3. The answer is strong advice.<br><br>CONTEXT: Therefore, this regimen would be a viable option for acne treatments either as monotherapy or as combination therapy.<br>QUESTION: Is this a (0) no advice, (1) weak advice, or (2) strong advice? Let's think step by step.<br>ANSWER:<br>1. The term "would be a viable option" indicates that there is an advice in the context.<br>2. "Would be" indicates that the opinion is not strong.<br>3. The answer is weak advice.<br><br>CONTEXT: {context}<br>QUESTION: Is this a (0) no advice, (1) weak advice, or (2) strong advice? Let's think step by step. | Classification: Advice in structured abstracts | Few-shot CoT |
| "CONTEXT: The high rate of text message usage makes it feasible to recruit YAMs for a prospective study in which personalized text messages are used to promote healthy behaviors.<br>QUESTION: Does 2 - correlation, 3 - conditional causation, or 4 – direct causation expressed in the sentence, or it is a 1 - no relationship sentence?<br>Let's think step by step:<br>1) The sentence does not describe any correlation or causation relation between two entities.<br>2) It is a no relationship sentence.<br>Answer: 1 - no relationship<br><br>CONTEXT: The incidence of falls and poor quality of life may be partially associated with the presence of depression.<br>QUESTION: Does 2 - correlation, 3 - conditional causation, or 4 – direct causation expressed in the sentence, or it is a 1 - no relationship sentence?<br>Let's think step by step:<br>1) The sentence describes the relationship between "the incidence of falls and poor quality of life" and "depression".<br>2) The relationship between them is correlation, as the authors used the language cue "associated with" to describe it. "Associated with" is a commonly used expression to indicate correlation.<br>3) Therefore, the sentence expresses correlation.<br>Answer: 2 - correlation<br><br>CONTEXT: Our study provides preliminary evidence that mothers who consume diets higher in fruit and lower in fried foods and cured meats during pregnancy may reduce the risk of unilateral retinoblastoma in their offspring.<br>QUESTION: Does 2 - correlation, 3 - conditional causation, or 4 – direct causation expressed in the sentence, or it is a 1 - no relationship sentence?<br>Let's think step by step:<br>1) The sentence describes the relationship between "diets higher in fruit and lower in fried foods and cured meats" and "the risk of "unilateral retinoblastoma in offspring".<br>2) The authors used the verb "reduce", which suggests that there is a direct effect between them.<br>3) When describing the effect, the authors also used a hedge word "may" to tone down the certainty of the effect.<br>3) Therefore, the sentence expresses conditional causation only.<br>Answer: 3 - conditional causation<br><br>CONTEXT: The nutritional course for patients undergoing colon surgery can be improved by implementing early oral nutritional supplements in the PACU.<br>QUESTION: Does 2 - correlation, 3 - conditional causation, or 4 – direct causation expressed in the sentence, or it is a 1 - no relationship sentence?<br>Let's think step by step:<br>1) The sentence describes the relationship between "nutritional course" and "the risk of "oral nutritional supplements in the PACU".<br>2) The authors used the verb "can be improved by", which suggests that there is a direct effect between them.<br>3) When describing the effect, the authors also used a modal verb "can", suggesting that the certainty of the effect is strong and direct. And no hedge word is used to tone down the certainty.<br>3) Therefore, the sentence expresses causation.<br>Answer: 4 - causation | Reasoning:<br>Causal relation detection | Few-shot CoT |

*Table B2: Prompt templates for Chain-of-Thought (CoT)*